\documentclass[10pt,twocolumn,letterpaper]{article}

\usepackage{cvpr}              
\usepackage{tabularx}
\usepackage{bbding}
\usepackage{multirow}
\definecolor{cvprblue}{rgb}{0.21,0.49,0.74}
\usepackage[pagebackref,breaklinks,colorlinks,allcolors=cvprblue]{hyperref}

\def\paperID{} 
\def\confName{CVPR}
\def\confYear{2026}

\title{VIMCAN: Visual-Inertial 3D Human Pose Estimation \\with Hybrid Mamba-Cross-Attention Network}

\author{
Zepeng Yang$^{1,2\ast}$\hspace{4mm} 
Junxuan Bai$^{3\ast}$\hspace{4mm} 
Hao Li$^{1,2}$\hspace{4mm} 
Ju Dai$^{2\dagger}$\hspace{4mm} 
Junjun Pan$^{1,2}$\hspace{4mm} 
Yongfeng Yin$^{1\dagger}$\hspace{4mm} 
Bin Li$^{4}$\hspace{4mm}\vspace{+1mm}\\
$^1$Beihang University \hspace{4mm}
$^2$Peng Cheng Laboratory \\
$^3$Capital University of Physical Education and Sports\\
$^4$Shenzhen Institutes of Advanced Technology, Chinese Academy of Sciences\\
{}
\vspace{-12mm}
}

\begin{document}
\maketitle
\begin{abstract}
\vspace{-0.4cm}

The rapid advances in deep learning have significantly enhanced the accuracy of multimodal 3D human pose estimation (HPE). However, the state-of-the-art (SOTA) HPE pipelines still rely on Transformers, whose quadratic complexity makes real-time processing for long sequences impractical. Mamba addresses this issue through selective state-space modeling, enabling efficient sequence processing without sacrificing representational power. Nevertheless, it struggles to capture complex spatial dependencies in multimodal settings. To bridge this gap, we propose VIMCAN, a hybrid architecture that combines the efficient sequence modeling of Mamba with the spatial reasoning of Cross-Attention, and performs robust visual–inertial fusion and human pose estimation between RGB keypoints and wearable IMU data. By leveraging Mamba’s dynamic parameterization for temporal modeling and Attention for spatial dependency extraction, VIMCAN achieves superior accuracy, with mean per-joint position errors (MPJPE) of 17.2 mm on TotalCapture and 45.3 mm on 3DPW. VIMCAN outperforms prior Transformer-based and other SOTA approaches while supporting real-time inference at over 60 frames per second on consumer-grade hardware. The source code is available on https://github.com/Eddieyzp/VIMCAN.
\end{abstract}

\footnotetext[1]{$\ast$ Zepeng Yang and Junxuan Bai contributed equally to this work.}
\footnotetext[2]{$\dagger$ Corresponding authors: daij@pcl.ac.cn and yyf@buaa.edu.cn}
\section{Introduction}
\label{sec:intro}

\begin{figure}[t]
  \centering
  
   \includegraphics[width=1\linewidth]{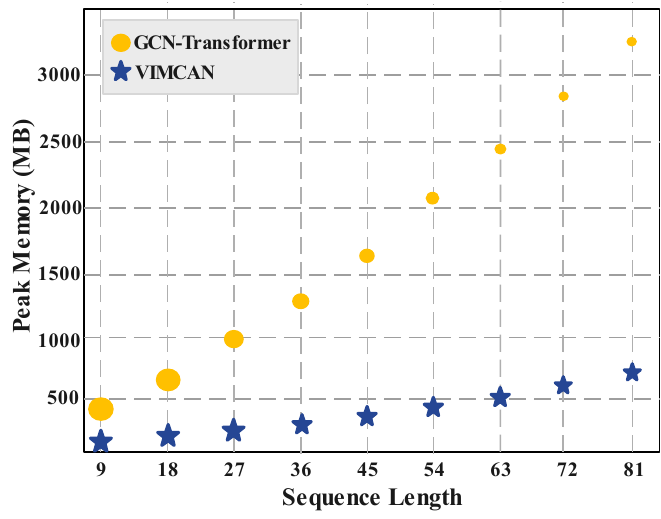}
   \caption{The comparison of peak memory usage during inference for a GCN-Transformer-based model~\cite{wangDeepLearningenabledVisualinertial2025} and the proposed VIMCAN. The peak memory usage at different lengths of input sequence. The x-axis denotes sequence length, and the y-axis represents peak memory usage (lower is better, in MB). The circles indicate the GCN-Transformer-based model, while the stars denote VIMCAN. The size of Symbol reflects the memory I/O throughput, illustrating computational efficiency during on-device inference.}
   \label{fig:motivation}
   \vspace{-0.2cm}
\end{figure}

3D human pose estimation (HPE) is a noteworthy research topic in computer vision, encompassing both vision-based and multimodal approaches. In recent years, the development of deep neural networks has significantly improved the accuracy of vision-based HPE, with prominent examples including Transformer~\cite{vaswaniAttentionAllYou2017a, zheng3DHumanPose2021, liMHFormerMultihypothesisTransformer2022, zhaoPoseFormerV2ExploringFrequency2023, zhuMotionBERTUnifiedPerspective2023, mehrabanMotionAGFormerEnhancing3D2024a}, Graph Convolutional Network (GCN)~\cite{zhaoSemanticGraphConvolutional2019, yuGLAGCNGloballocalAdaptive2023}, and Temporal Convolutional Network (TCN)~\cite{pavllo3DHumanPose2019, liuAttentionMechanismExploits2020}. However, vision-based HPE still faces two critical issues: (1) depth ambiguity when lifting 2D keypoints to 3D poses, and (2) inefficient computation when processing long sequences. To address the depth ambiguity problem, researchers have explored multimodal approaches by incorporating additional modalities alongside RGB images, such as depth data~\cite{shahroudyNTURGB+DLarge2016, liuNTURGB+D1202020, liHighqualityIndoorScene2022a}, inertial measurement unit (IMU) data~\cite{jiangTransformerInertialPoser2022, yiTransPoseRealtime3D2021, huangDeepInertialPoser2018,vonmarcardSparseInertialPoser2017a,liVisualInertialFusionbased2023}, and Light Detection and Ranging (LiDAR) data~\cite{yanRELI11DComprehensiveMultimodal2024a, anPretrainingDensityawarePose2025}. Thanks to its strong representation power, Transformer~\cite{vaswaniAttentionAllYou2017a} has become a cornerstone for multimodal learning by leveraging Cross-Attention mechanism. 
Nevertheless, as shown in Fig.~\ref{fig:motivation}, the quadratic complexity $\mathcal{O}(L^2)$ of Attention with respect to sequence length $L$, leads to high computational and memory demands, thereby limiting the scalability and applicability of these models in resource-constrained environments and high-throughput scenarios.

\begin{figure*}[!h]
    \centering
    \includegraphics[width=1\linewidth]{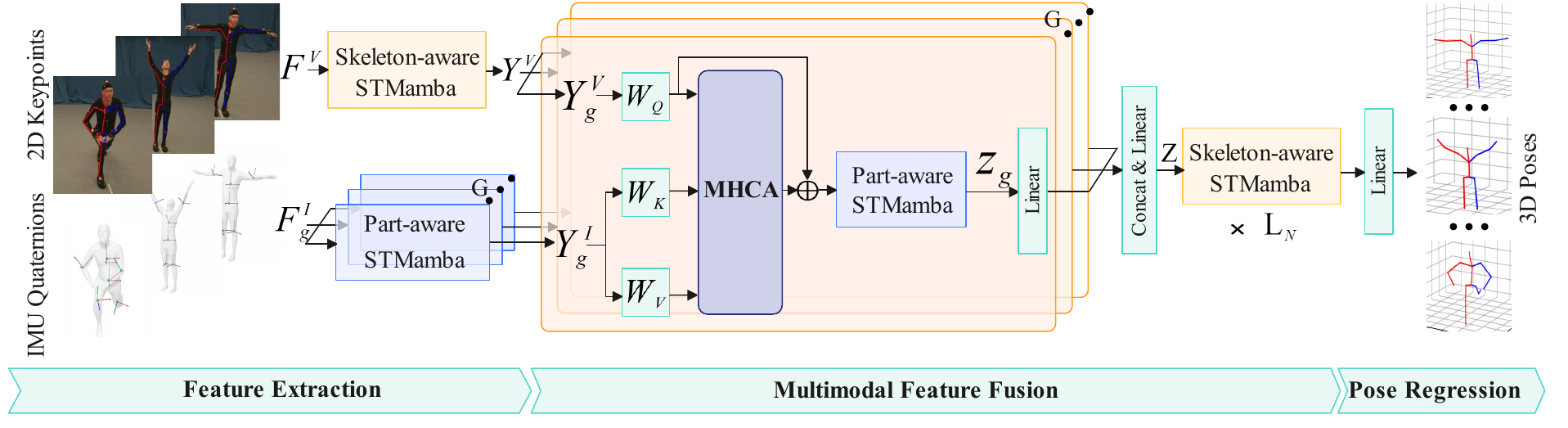}
    \caption{The framework of VIMCAN.}
    \label{fig:arch}
    \vspace{-0.3cm}
\end{figure*}

Mamba~\cite{guMambaLineartimeSequence2024,daoTransformersAreSsms2024} addresses the efficiency bottleneck by casting sequential data as a time-varying system and compressing historical context into a dynamically evolving hidden state, enabling recursive updates with linear complexity $\mathcal{O}(L)$. This feature facilitates efficient processing for long sequences, positioning Mamba as an alternative to Transformers. 
However, its spatial-reasoning capacity still lags behind that of convolution- or Attention-based approaches~\cite{yuMambaOutWeReally2025}, limiting its utility in multimodal tasks that require precise spatial modeling.

To overcome the above limitations, we propose VIMCAN, a \textbf{V}isual-\textbf{I}nertial \textbf{M}amba-\textbf{C}ross-\textbf{A}ttention \textbf{N}etwork for human pose estimation. 
The hybrid design takes RGB keypoints and IMU data as input, leveraging Mamba's efficient sequence processing and Cross-Attention's aptitude for cross-modal fusion to achieve superior performance.
Our approach strikes a balance between accuracy, robustness, and efficiency, enabling real-time inference exceeding 60 frames per second on consumer-grade hardware. 
Extensive experiments on TotalCapture and 3DPW validate the superiority of VIMCAN, with MPJPE reaching 17.2 mm and 45.3 mm, respectively, outperforming previous Transformer-based methods and other state-of-the-art (SOTA) approaches.

The main contributions include:
\begin{itemize}
\item We propose VIMCAN, a Mamba-Cross-Attention architecture that combines Mamba's efficient sequence processing with Cross-Attention's robust multimodal fusion, enabling real-time variable-length inference.

\item To the best of our knowledge, VIMCAN is the first work to apply Mamba to vision-inertial multimodal 3D human pose estimation, addressing depth ambiguity and computational complexity.
\item Extensive experiments demonstrate the superior accuracy, robustness, and efficiency of VIMCAN over SOTA methods.
\end{itemize}

\section{Related Work}
\subsection{RGB-based 3D HPE}
3D human pose estimation from RGB images or videos has been revolutionized by novel deep architectures, including GCN~\cite{zhaoSemanticGraphConvolutional2019,yuGLAGCNGloballocalAdaptive2023}, TCN~\cite{pavllo3DHumanPose2019,liuAttentionMechanismExploits2020}, and Transformer~\cite{vaswaniAttentionAllYou2017a,zheng3DHumanPose2021,liMHFormerMultihypothesisTransformer2022,zhaoPoseFormerV2ExploringFrequency2023,zhuMotionBERTUnifiedPerspective2023,mehrabanMotionAGFormerEnhancing3D2024a}. GCN-based models effectively capture spatial dependencies and temporal dynamics for motion sequences on a skeletal graph, while TCN-based approaches utilize strided convolutions to extract long-range dependencies. Transformer-based networks, on the other hand, exploit attention mechanisms to learn global context relationships. For instance, PoseFormer~\cite{zheng3DHumanPose2021} employs separate Transformer modules to encode intra-frame joint relations and inter-frame temporal dependencies. 
Similarly, KD-Former~\cite{KDformerKinematicDynamic2023} divides the human body into kinematic groups and integrates kinematic and dynamic features via a non-autoregressive Transformer, enhancing overall representational capacity. 
These advances underscore the pivotal role of attention mechanisms and structural priors in achieving superior accuracy and generalization. Nevertheless, uni-modal approaches are inherently limited by depth ambiguity in the 2D-to-3D lifting process.

\subsection{Multimodal 3D HPE}
Recovering 3D human pose from 2D observations is inherently ill-posed. 
Multimodal fusion methods with different sensors have enhanced the accuracy and robustness of 3D HPE, including RGB-D cameras~\cite{shahroudyNTURGB+DLarge2016,liuNTURGB+D1202020,liHighqualityIndoorScene2022a}, IMUs~\cite{jiangTransformerInertialPoser2022, yiTransPoseRealtime3D2021, huangDeepInertialPoser2018,vonmarcardSparseInertialPoser2017a}, and LiDAR data~\cite{yanRELI11DComprehensiveMultimodal2024a,anPretrainingDensityawarePose2025}. Such fusion exploits the complementary properties of sensors to overcome the limitations of single modalities. For instance, RGB cameras provide detailed visual context, while IMUs deliver high-frequency, low-latency dynamic data, thereby overcoming occlusion and blur issues inherent in visual systems~\cite{liVisualInertialFusionbased2023,zhaoReviewWearableIMU2018}. When performing multimodal fusion, Cross-Attention is frequently employed due to its ability to model intricate relationships between heterogeneous inputs~\cite{xuMultimodalLearningTransformers2023}. 
Wang~\etal~\cite{wangDeepLearningenabledVisualinertial2025} propose a GCN-Transformer hybrid architecture with Cross-Attention to fuse visual and IMU data, using kinematic priors for modality grouping and robust integration. Similarly, Liu~\etal~\cite{liu3DHumanPose2024} utilize a pre-trained CNN to extract 2D image features as keys and values, with IMU features as queries, to regress Skinned Multi-Person Linear (SMPL)~\cite{loperSMPLSkinnedMultiperson2015} parameters and root joint translation. 
However, their studies found that to achieve real-time inference with Attention, the temporal window length must be constrained.
Limitations in computational efficiency, memory usage, and hardware compatibility hinder its application in scenarios involving long sequences and on resource-constrained devices.

\subsection{Mamba Architectures}
Mamba~\cite{guMambaLineartimeSequence2024} emerges as a promising alternative to Transformer architectures, due to its linear time complexity in sequence modeling. Its efficacy in language processing has motivated its extension to visual representation learning. 
Zhu~\etal~\cite{zhuVisionMambaEfficient2024} introduce Vision Mamba, which integrates bidirectional State Space Models (SSMs) with positional embeddings to overcome limitations in spatial modeling. 
Liu~\etal~\cite{liuVMambaVisualState2024} propose VMamba, featuring efficient spatio-temporal context encoding via a Cross-Scan module. 
Hu~\etal~\cite{huExploitingMultimodalSpatialtemporal2024} develope a multimodal video object tracking framework that leverages Mamba's temporal modeling via a dedicated temporal state generator, enabling cross-modal spatio-temporal feature extraction. 
In 3D HPE, Huang~\etal~\cite{huangPoseMambaMonocular3D2025} adopt a global-local spatio-temporal scanning mechanism with linear complexity to exploit human skeletal geometry. 
Moreover, hybrid architectures combining Mamba with structural priors have been developed. 
Zhang~\etal~\cite{zhangPoseMagicEfficient2025} introduce Pose Magic, a Mamba-GCN model which enhances local joint relationships while maintaining efficient long-range modeling and supports both bidirectional and causal reasoning. 
Despite these advances, an empirical study by Yu~\etal~\cite{yuMambaOutWeReally2025} reveals that pure Mamba falls short of convolutional or attention-based methods in capturing intricate spatial relationships. To mitigate this, Hatamizadeh~\etal~\cite{hatamizadehMambaVisionHybridMambatransformer2025} propose a Mamba-Transformer hybrid architecture that combines the efficiency of Mamba with the long-range dependency modeling capability of Attention, thereby establishing a benchmark for efficient multimodal 3D HPE.

\section{Methodology}
\label{sec:method}
\begin{figure}[t]
    \centering
    \includegraphics[width=1\linewidth]{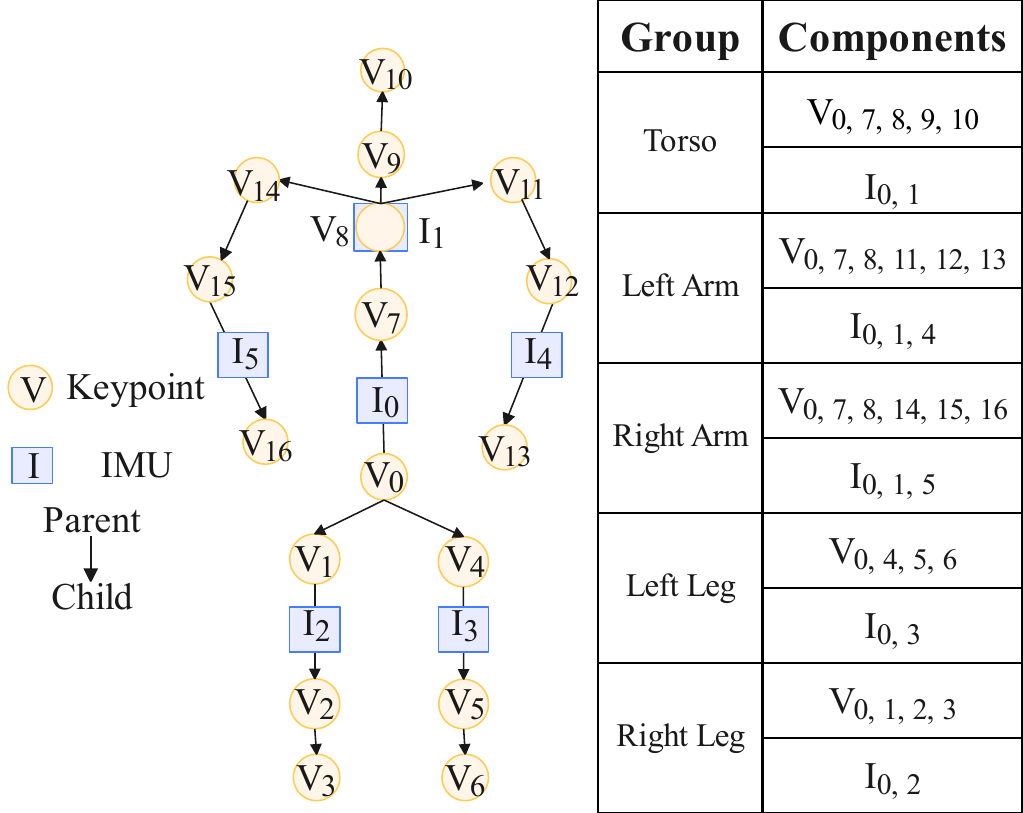}
    \caption{Illustration of skeleton topology and group components.}
    \label{fig:skel}
    \vspace{-0.5cm}
\end{figure}

\subsection{Feature Extraction}
As shown in Fig.~\ref{fig:arch}, during feature extraction, VIMCAN learns features from visual images and IMU inputs. Specifically, VIMCAN receives $J$=17 keypoints coordinates from images and unit quaternions from $I$=6 IMUs as inputs. 
Drawing from kinematic priors~\cite{KDformerKinematicDynamic2023,wangDeepLearningenabledVisualinertial2025}, the IMUs are partitioned into $G$=5 groups (torso, left arm, right arm, left leg, and right leg, shown in Fig.~\ref{fig:skel}), each containing $I_g$ IMUs.
Both modalities are first lifted to a common embedding dimension $D_{e}$ via linear layers, yielding visual feature $F^V \in \mathbb{R}^{T \times J \times D_{e}}$ and inertial features $F^I_g \in \mathbb{R}^{T \times I_g \times D_{e}}$.
To efficiently capture spatial and temporal information, each modality is processed independently using Spatio-Temporal Mamba (STMamba) modules. Specifically, $F^V$ is processed via skeleton-aware STMamba, while $F^I_g$ from each body part $g$ is handled by dedicated part-aware STMamba.

\begin{figure}[!h]
    \centering
    \includegraphics[width=1\linewidth]{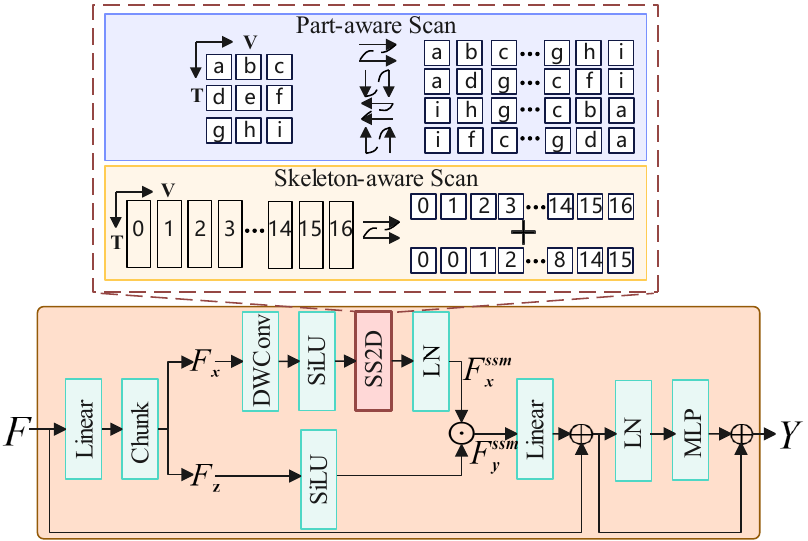}
    \caption{The architecture of BiSTSSM module.}
    \label{fig:ssm}
    \vspace{-0.5cm}
\end{figure}

As shown in Fig.~\ref{fig:ssm}, each STMamba comprises two Bidirectional Spatio-Temporal State Space Model (BiSTSSM) blocks. The input is first processed spatially by a BiSTSSM to encode intra-frame topological relationships, followed by a temporal BiSTSSM to model inter-frame dynamics. The primary distinction between part-aware and skeleton-aware STMamba lies in the BiSTSSM scanning mechanism, tailored to each modality.

For the inertial branch, a bidirectional spatio-temporal selective scan (SS2D)~\cite{liuVMambaVisualState2024} aggregates information across four directions: spatial forward/reverse and temporal forward/reverse. Regarding the visual branch, inspired by~\cite{huangPoseMambaMonocular3D2025}, keypoints are reordered according to parent-child joint relationships (defined in Fig.~\ref{fig:skel}) during spatial forward scanning to incorporate kinematic priors.

Prior to spatial feature extraction, $F^V$ and $F^I_{g}$ are augmented with learnable spatial positional encodings and layer normalization (LN). 
Within each BiSTSSM, the input features $F$ (either $F^V$ or $F^I_g$) are projected to an internal SSM dimension using a linear layer (FC) to facilitate modeling of complex spatio-temporal dependencies. The projected tensor is then divided (Chunk) into two components:
\begin{equation}
\setlength\abovedisplayskip{3pt}
\setlength\belowdisplayskip{3pt}
    F_{x}, F_{z} = \text{Chunk}(\text{FC}(F)).
\end{equation}

For $F_{x}$, a Depthwise Convolution (DWConv) layer is leveraged to capture local patterns, followed by an activation function (SiLU) to enhance local feature interactions. The scanning yields an intermediate tensor:
\begin{equation}
\setlength\abovedisplayskip{3pt}
\setlength\belowdisplayskip{3pt}
    F_{x}^{ssm} = \text{LN}(\text{SS2D}(\sigma(\text{DWConv}(F_{x})))).
\end{equation}

Concurrently, $F_z$ serves as a gating signal. The gated output is computed as:
\begin{equation}
\setlength\abovedisplayskip{3pt}
\setlength\belowdisplayskip{3pt}
    F_{y}^{ssm} = F_{x}^{ssm} \cdot \sigma(F_{z}),
\end{equation}
which is then projected back to $D_{e}$, combined with a residual connection, and processed by an MLP:
\begin{equation}
\setlength\abovedisplayskip{3pt}
\setlength\belowdisplayskip{3pt}
    \begin{split}
    &Y^{ssm} = \text{FC}(F_{y}^{ssm}) + F, \\
    &Y = \text{MLP}(\text{LN}(Y^{ssm})) + Y^{ssm}.
    \end{split}
\end{equation}
Here, $Y$ denotes the extracted visual feature $Y^V$ or inertial feature $Y_g^I$ for group $g$.

Subsequently, to model joint-wise spatio-temporal correlations within sliding windows, temporal positional encodings are added to $Y$, followed by a temporal BiSTSSM block that comprehensively captures inter-frame dependencies. The temporal scanning mechanism aligns with the spatial one, adapted per modality.

\subsection{Multimodal Feature Fusion}
Although Mamba's state-space model attains linear complexity $\mathcal{O}(L)$, its selective mechanism compromises its expressive capacity for efficiency, resulting in information loss that hinders cross-modal interactions in tasks that demand precise spatial reasoning~\cite{yuMambaOutWeReally2025}.
Conversely, Attention mechanisms are utilized to aggregate information across all tokens, rendering cross-attention well-suited for capturing intricate cross-modal spatial dependencies~\cite{xuMultimodalLearningTransformers2023}.

To facilitate robust fusion of visual and inertial features, we introduce a Cross-Attention module as the backbone. The architecture of the module is depicted in the appendix.

The extracted visual features $Y^V$ are partitioned into groups $Y^V_g$, each containing $J_g$ joints. In the Cross-Attention module, Multi-Head Cross-Attention (MHCA) integrates the multimodal inputs. For each group $g$, visual features serve as queries $Q^V_g$, while inertial features yield keys $K^I_g$ and values $V^I_g$. The outputs from each head $h$ are concatenated and normalized, with a residual connection applied solely to the visual queries to retain skeletal information:
\begin{equation}
\setlength\abovedisplayskip{3pt}
\setlength\belowdisplayskip{3pt}
    \begin{split}
        \text{MHCA} &= \text{Concat}\left[\text{Softmax}\left(\frac{Q_g^V {K_g^I}^\top}{\sqrt{d_k}}\right) V_g^I\right]_h, \\
        Z_g &= \text{LN}(\text{MHCA}) + Q^V_g,
    \end{split}
\end{equation}
where $d_k$ denotes the key dimension.

As illustrated in Fig.~\ref{fig:arch}, the initial fused features $Z_g \in \mathbb{R}^{T \times (J_g \cdot D_{e})}$ for each group are obtained via a part-aware STMamba module. A fully connected (FC) layer then reduces the feature dimension from $J_g \cdot D_{e}$ to $D_g$. The resulting group-wise fusion features are concatenated to construct the global fused representation $Z \in \mathbb{R}^{T \times (G \cdot D_g)}$. Subsequently, another FC layer transforms the feature dimension per frame from $G \cdot D_{g}$ to $J \cdot D_g$. Finally, a stack of $L_N$ skeleton-aware STMamba performs global modeling to capture comprehensive spatio-temporal dependencies.

\begin{figure}[!t]
    \centering
    \includegraphics[width=0.7\linewidth]{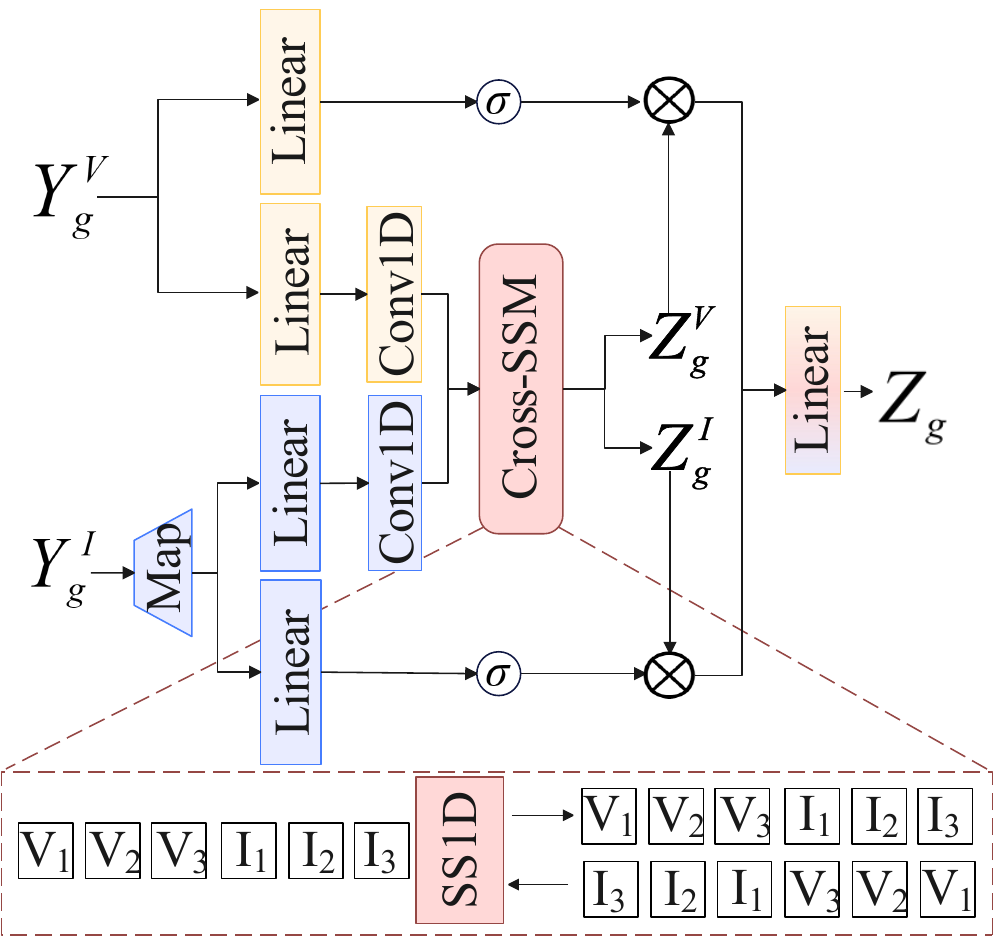}
    \caption{The architecture of Cross-Mamba module.}
    \vspace{-0.5cm}
    \label{fig:cmamba}
\end{figure}

\subsection{Cross-Mamba: An Alternative for Fusion}
To verify that our design overcomes Mamba’s limitations, we conduct a comparative evaluation.
We further propose a Cross-Mamba module as an alternative, which is illustrated in Fig.~\ref{fig:cmamba}. 
A linear layer adaptively maps inertial features $Y^I_g$ to visual features $Y^V_g$, with weights learned during training to capture inter-modal significance. Both modalities undergo linear projections and 1D convolutions to refine local patterns. Deviating from the four-directional scanning in SS2D, the Cross State Space Model (Cross-SSM) concatenates visual and inertial features along the spatial axis and applies bidirectional scanning (SS1D) for alignment with cross-attention. The resulting fused features are segmented into visual $z^V_g$ and inertial $z^I_g$ components, modulated via activated mapped features (SiLU), and then concatenated and projected to a unified group embedding dimension $D_g$, producing $Z_g$.

\subsection{Pose Regression}
The regression generates root-relative 3D poses for the entire sequence via layer normalization and a linear layer, transforming fused multimodal features into output pose $ P \in \mathbb{R}^{T \times J \times 3} $  (3D coordinates of $ J $ joints across $ T $ frames).

\subsection{Loss Function}
To train VIMCAN, we utilize a composite loss function to promote accurate 3D pose estimation while enforcing temporal coherence. The loss function comprises four components: mean per-joint position error (MPJPE), normalized MPJPE (N-MPJPE), mean per-joint velocity error (MPJVE), and temporal consistency (TC) loss.

The MPJPE loss ($\mathcal{L}_{\text{MPJPE}}$) quantifies the L2 distance between predicted 3D poses $\mathbf{\hat{J}}$ and ground-truth (GT) coordinates $\mathbf{J}$ across batches $B$, time steps $T$, and joints $J$:
\begin{equation}
\setlength\abovedisplayskip{3pt}
\setlength\belowdisplayskip{3pt}
    \mathcal{L}_{\text{MPJPE}} = \frac{1}{B \cdot T \cdot J} \sum_{b=1}^B \sum_{t=1}^{T} \sum_{j=1}^{J} \| \mathbf{J}_{b,t,j} - \mathbf{\hat{J}}_{b,t,j} \|^2.
\end{equation} 

To mitigate potential global-scale drift between the predicted and ground-truth sequences, we first use the least-squares method to compute a scale factor $s$, and then compute the normalized MPJPE loss ($\mathcal{L}_{\text{N-MPJPE}}$) between the scaled prediction and the target:
\begin{equation}
\setlength\abovedisplayskip{3pt}
\setlength\belowdisplayskip{3pt}
    \mathcal{L}_{\text{N-MPJPE}} = \frac{1}{B \cdot T \cdot J} \sum_{b=1}^B \sum_{t=1}^{T} \sum_{j=1}^{J} \| s \mathbf{J}_{b,t,j} - \mathbf{\hat{J}}_{b,t,j} \|^2.
\end{equation} 

For temporal smoothness, the MPJVE loss~\cite{pavllo3DHumanPose2019} ($\mathcal{L}_{\text{V}}$) measures the L2 distance between the first-order derivatives of predicted and GT joint sequences, ensuring consistent motion velocities across frames:
\begin{equation}
\setlength\abovedisplayskip{3pt}
\setlength\belowdisplayskip{3pt}
    \begin{split}
        \text{PJVE} &= \| (\mathbf{J}_{b,t+1,j} - \mathbf{J}_{b,t,j}) - (\mathbf{\hat{J}}_{b,t+1,j} - \mathbf{\hat{J}}_{b,t,j}) \|^2, \\
        \mathcal{L}_{\text{V}}& = \frac{1}{B\cdot (T-1)\cdot J} \sum_{b=1}^{B}\sum_{t=1}^{T-1}\sum_{j=1}^{J} \text{PJVE}.
    \end{split}
\end{equation}

Additionally, the TC loss~\cite{hossainExploitingTemporalInformation2018} ($\mathcal{L}_{\text{TC}}$) imposes a weighted L2 penalty on first-order differences in predicted poses to suppress abrupt fluctuations:
\begin{equation}
\setlength\abovedisplayskip{3pt}
\setlength\belowdisplayskip{3pt}
    \mathcal{L}_{\text{TC}} = \frac{1}{B \cdot (T-1) \cdot J}\sum_{b=1}^{B}\sum_{t=1}^{T-1}\sum_{j=1}^{J} w_{j}\bigl\| \hat{\mathbf{J}}_{b,t+1,j} - \hat{\mathbf{J}}_{b,t,j} \bigr\|^{2},
\end{equation}
where $w_j$ assigns higher penalties to perceptually critical joints, such as distal limbs.

The total loss is a weighted sum of these terms, enabling end-to-end training:
\begin{equation}
\setlength\abovedisplayskip{3pt}
\setlength\belowdisplayskip{3pt}
\begin{split}
    \mathcal{L}_{\text{Total}} &= \lambda_{\text{MPJPE}} \cdot \mathcal{L}_{\text{MPJPE}} 
    + \lambda_{\text{N-MPJPE}} \cdot \mathcal{L}_{\text{N-MPJPE}} \\
    &+ \lambda_{\text{V}} \cdot \mathcal{L}_{\text{V}}
    + \lambda_{\text{TC}} \cdot \mathcal{L}_{\text{TC}},
\end{split}
\end{equation}
where $ \lambda_{\text{MPJPE}} $, $ \lambda_{\text{N-MPJPE}} $, $ \lambda_{\text{V}} $, and $ \lambda_{\text{TC}} $ are weighting coefficients optimized during training.
Detailed parameters can be found in Sec.~\ref{sec:exp}.

\subsection{Training strategy}
VIMCAN inherently supports inference on variable-length sequences, obviating the need for padding, masking, or fixed-length constraints prevalent in conventional approaches.
To utilize this flexibility, we set a maximum sequence length of T = 81 and randomly sample sequences of lengths \{9, 18, 27, 36, 45, 54, 63, 72, 81\} within each training batch.

\begin{table*}[!ht]
  \centering
  \caption{The performance comparison on TotalCapture dataset with different visual-inertial fusion methods. \#IMUs: Number of IMUs used. 2D: Type of 2D pose detector (MP: MediaPipe, SN: SimpleNet, GT: Ground-Truth). S: Seen subject. U: Unseen subject. W: Walking. A: Acting. FS: Freestyle. P1: Average MPJPE (mm) across all test sets (lower is better). P2: Procrustes-aligned MPJPE (mm, lower is better). } 
    \begin{tabular}{ccccccccccc}
      \hline
      Method & \#IMUs & 2D & SW2 & SFS3 & SA3 & UW2 & UFS3 & UA3 & P1 $\downarrow$ & P2 $\downarrow$\\
      \hline
        Pan's~\cite{panFusingMonocularImages2023}
        & 6 & MP & 39.3 & 62.0 & 45.6 & 38.2 & \underline{\textbf{43.6}} & 62.1 & 48.1 & 32.9 \\ 
        Wang's~\cite{wangDeepLearningenabledVisualinertial2025}
        & 6 & MP & \underline{\textbf{17.7}} & 48.3 & 29.1 & 39.6 & 64.1 & 50.2 & 39.0 & 28.8 \\ 
        VIMCAN (Ours)
        & 6 & MP & 18.5 & \underline{\textbf{39.8}} & \underline{\textbf{24.1}} & \underline{\textbf{34.8}} & 58.6 & \underline{\textbf{44.3}} & \underline{\textbf{33.2}} & \underline{\textbf{25.7}} \\ 
        \hline
        Wang's~\cite{wangDeepLearningenabledVisualinertial2025}
        & 6 & SN & \underline{\textbf{16.4}} & 41.7 & 25.2 & 35.4 & 62.0 & 44.8 & 34.9 & 26.9 \\
        VIMCAN (Ours)
        & 6 & SN & 17.6 & \underline{\textbf{39.1}} & \underline{\textbf{22.3}} & \underline{\textbf{33.8}} & \underline{\textbf{58.7}} & \underline{\textbf{42.6}} & \underline{\textbf{31.2}} & \underline{\textbf{23.6}} \\
        \hline
        Bao's~\cite{baoHybrid3DHuman2024} 
        & 8 & SN & \underline{\textbf{15.7}} & 47.3 & 20.1 & \underline{\textbf{24.5}} & 60.2 & 50.5 & 33.7 & - \\
        Wang's~\cite{wangDeepLearningenabledVisualinertial2025}
        & 8 & SN & 17.8 & 40.4 & 23.2 & 33.7 & 58.4 & 41.6 & 33.4 & 25.1\\
        VIMCAN (Ours)
        & 8 & SN & 17.0 & \underline{\textbf{33.8}}& \underline{\textbf{19.6}} & 33.2 & \underline{\textbf{56.5}} & \underline{\textbf{40.5}} & \underline{\textbf{28.9}} & \underline{\textbf{21.3}} \\ 
        \hline
        Wang's~\cite{wangDeepLearningenabledVisualinertial2025}
        & 6 & GT & 12.9 & 29.7 & 18.7 & 37.2 & 46.9 & 41.9 & 28.6 & 17.6 \\ 
        VIMCAN (Ours)
        & 6 & GT & \underline{\textbf{8.0}}  & \underline{\textbf{23.7}} & \underline{\textbf{14.2}} & \underline{\textbf{25.7}} & \underline{\textbf{42.3}} & \underline{\textbf{36.3}} & \underline{\textbf{17.2} }& \underline{\textbf{13.8}} \\ 
      \hline
    \end{tabular}    
\label{tab:tc-comparison}
\vspace{-0.2cm}
\end{table*}

\section{Experiments}
\label{sec:exp}
\subsection{Datasets and Evaluation Metrics}
We assess VIMCAN on two benchmark datasets for 3D HPE: TotalCapture~\cite{trumbleTotalCapture3D2017} and 3DPW~\cite{marcardRecoveringAccurate3D2018}.
The TotalCapture dataset consists of 1.9 million frames with synchronized multi-view videos from eight cameras, data from 13 IMUs on body segments, and Vicon-derived ground-truth annotations, making it suitable for visual-inertial fusion evaluation. It encompasses five subjects performing four actions—Range of Motion (ROM), Walking, Acting, and Freestyle—each in three trials. Consistent with prior protocols, the train set includes ROM (1, 2, 3), Walking (1, 3), Acting (1, 2), and Freestyle (1, 2) from subjects 1, 2, and 3. The test set comprises Walking (2), Acting (3), and Freestyle (3) across all subjects, facilitating evaluation of generalization to both seen and unseen subjects.

Similarly, the 3DPW dataset captures multiple actors engaging in diverse natural activities and contains various actions, such as playing basketball, walking, and freestyling. 
The images are sampled at 30 frames per second, and the poses are collected at 60 HZ, with SMPL parameters and 3D joints location ground truths. 
Due to the unavailability of direct IMU data, we generate synthetic IMU readings following~\cite{panFusingMonocularImages2023} and adopt the official train-valid-test split~\cite{marcardRecoveringAccurate3D2018}.

Performance is evaluated using MPJPE (denoted as P1) in millimeters for absolute positional accuracy and Procrustes-aligned MPJPE (P-MPJPE, denoted as P2), where the estimated poses are aligned with GT through a rigid transformation.
Memory I/O determines the computational efficiency of the model on the device during inference~\cite{leeEfficientViMEfficientVision2025}. We measure throughput in frames per second (FPS), peak memory usage, and parameter count. 

\subsection{Experiments Setup}
The VIMCAN model is implemented in PyTorch and trained on a single NVIDIA RTX 3090 GPU. Training is conducted with a batch size of 16 over 20 epochs, utilizing the AdamW optimizer with a weight decay of 0.01. The learning rate is initialized at $2 \times 10^{-4}$ and decays exponentially with a decay factor of 0.99. The loss weights are set as $\lambda_{\text{MPJPE}}$ =1, $\lambda_{\text{N-MPJPE}}$=0.5, $\lambda_{\text{V}}$=20, and $\lambda_{\text{TC}}$=0.5.
Network hyper-parameters include an embedding dimension $D_{e}$=64, group dimension $D_{g}$=256. With a stack of $L_N$=5 skeleton-aware STMamba layers for global modeling. 

\subsection{Data Pre-processing}
Visual inputs comprise 2D keypoint coordinates for $J$ joints, extracted from monocular RGB videos and normalized to $[-1, 1]$ to ensure scale invariance across frames. Inertial inputs comprise unit-quaternion measurements from 6 IMUs mounted on the pelvis, sternum, and limbs, capturing rotations in the sensors' local coordinate systems.
To align these heterogeneous modalities spatially, we compute the bone orientation in the camera frame as follows:
\begin{equation}
\mathbf{R}_{B}^{C} = \mathbf{R}_{B}^{I} \cdot \mathbf{R}_{I}^{S} \cdot \mathbf{R}_{S}^{G} \cdot \mathbf{R}_{G}^{C},
\end{equation}
where $ \mathbf{R}_{B}^{I} $ denotes the relative rotation between the IMU and the corresponding skeletal bone, $ \mathbf{R}_{I}^{S} $ is the measured orientation of the IMU,  $ \mathbf{R}_{S}^{G} $ transforms the IMU frame to a global coordinate system (\textit{i.e.}, Vicon), and $ \mathbf{R}_{G}^{C} $ accounts for extrinsic camera parameters. The terms $ \mathbf{R}_{B}^{I} $ and $ \mathbf{R}_{S}^{G} $ are derived via standardized T-pose calibration \cite{trumbleTotalCapture3D2017,huangDeepInertialPoser2018}.
\begin{table}[!t]\small
  \centering
  \caption{The performance comparison on 3DPW testing set using MediaPipe 2D pose detector.} 
    \begin{tabular}{ccccc}
      \hline
        Methods & Liu's & Pan's & Wang's & Ours  \\ \hline
        P1 & 60.3 & 55.0 & 53.9 & \underline{\textbf{45.3}} \\ 
        Refs. & \cite{liu3DHumanPose2024} & \cite{panFusingMonocularImages2023} & \cite{wangDeepLearningenabledVisualinertial2025} & - \\ \hline
    \end{tabular}
\label{tab:pw-comparison}
\end{table}
\subsection{Performance Comparison on Accuracy}
We compare VIMCAN against the state-of-the-art visual-inertial fusion methods on TotalCapture and 3DPW datasets, including Pan's RNN-based approach~\cite{panFusingMonocularImages2023} and Bao's TCN-based method~\cite{baoHybrid3DHuman2024}, Liu's CNN-Transformer hybrid framework~\cite{liu3DHumanPose2024}, and Wang's GCN-Transformer hybrid architecture~\cite{wangDeepLearningenabledVisualinertial2025}.

\begin{figure*}[t]
    \centering
    \includegraphics[width=1\linewidth]{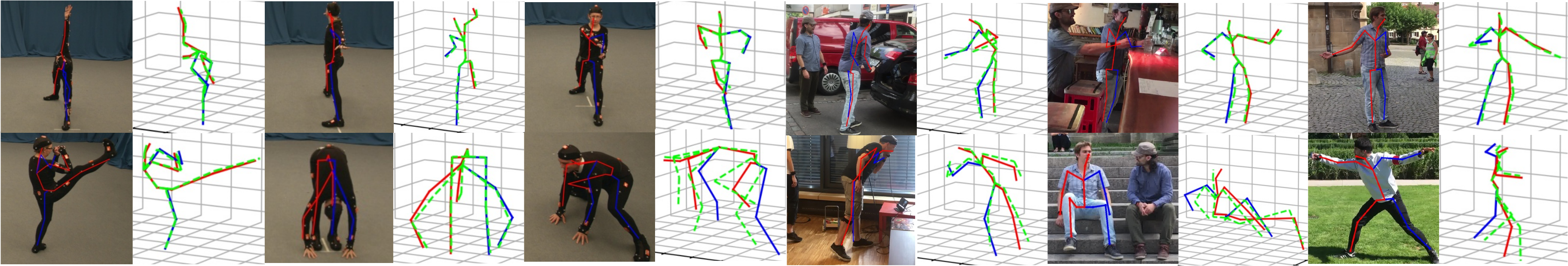}
    \caption{The qualitative analysis of VIMCAN. 
    The green dashed lines denote the ground truth, and other colored lines represent the predictions.
    }
    \vspace{-0.3cm}
    \label{fig:vis}
\end{figure*}

On TotalCapture, we simultaneously employ MediaPipe and SimpleNet as 2D pose detectors. The ground-truth 2D keypoints are also used to isolate the impact of 2D detection noises and evaluate upper-bound performance. 
Table~\ref{tab:tc-comparison} reports the P1 performance on subsets and the P1/P2 results on the entire dataset.
With 6 IMUs and MediaPipe, VIMCAN attains P1 of 33.2 mm and P2 of 25.7 mm, outperforming Pan's~\cite{panFusingMonocularImages2023} and Wang's~\cite{wangDeepLearningenabledVisualinertial2025}. 
Using SimpleNet with 6 IMUs, VIMCAN achieves P1 of 31.2 mm and P2 of 23.6 mm, exceeding the results of Wang's~\cite{wangDeepLearningenabledVisualinertial2025}. 
The correlation between the 2D detection quality and the final 3D pose estimation accuracy is evident, which is discussed in the appendix. 
For 8 IMUs with SimpleNet, where our configuration emphasizes the upper-body activity via additional upper-arm IMUs, VIMCAN yields P1 of 28.9 mm and P2 of 21.3 mm, surpassing the results of Bao's~\cite{baoHybrid3DHuman2024} and Wang's~\cite{wangDeepLearningenabledVisualinertial2025}. 
Although Bao's~\cite{baoHybrid3DHuman2024} performs better on Walking subset, VIMCAN exhibits stronger generalization to complex actions such as Freestyle 3 (FS3) and Acting 3 (A3) for unseen subjects. With the ground-truth keypoints and 6 IMUs, VIMCAN further reduces errors, improving over Wang's~\cite{wangDeepLearningenabledVisualinertial2025} by 11.4 mm on P1 and 3.8 mm on P2.
The qualitative result is visualized in Fig.~\ref{fig:vis}.

On 3DPW dataset, following Pan~\etal~\cite{panFusingMonocularImages2023} and Wang~\etal~\cite{wangDeepLearningenabledVisualinertial2025}, we use MediaPipe as the 2D pose detector and fine-tune VIMCAN on its training set. 
Specifically, the model was initially trained on TotalCapture dataset, and we fine-tuned the feature extractors and regressor. 
The learning rate was set to $1 \times 10^{-4}$ with the same decay strategy as above.
As presented in Table~\ref{tab:pw-comparison}, VIMCAN achieves 45.3 mm on P1, surpassing prior methods. 
These results highlight the robust accuracy and generalization capacity across diverse datasets and scenarios for VIMCAN.

\begin{table}[!t]\small
  \caption{The ablation study for variable-length training strategy using the ground-truth 2D inputs on TotalCapture testing set. T: Fixed lengths. V: Variable-length. P1: Average MPJPE (mm, lower is better).} 
    \begin{tabularx}{\linewidth}{ccccccXX}
      \hline
        T  & SW2 & SFS3 & SA3 & UW2 & UFS3 & UA3 & P1 $\downarrow$ \\ \hline
        9  & 9.7 & 26.6 & 14.9& 28.3& 44.7 &39.0 & 21.1 \\ 
        27 & 8.6 & 24.7 & 14.5& 26.4& 43.0 &37.7 & 19.2 \\ 
        81 & \underline{\textbf{8.0}} & \underline{\textbf{23.7}} & \underline{\textbf{14.2}} & \underline{\textbf{25.7}} & \underline{\textbf{42.3}} & \underline{\textbf{36.3}} & \underline{\textbf{17.2}} \\ 
        V  & 8.4 & 24.2 & 14.5& 26.4& 42.8 &37.3 & 18.9 \\ \hline
    \end{tabularx}
\label{tab:dynamic}
\end{table}
\begin{table}[!t]\small
  \centering
  \caption{The ablation study for grouping and skeleton-aware scanning on TotalCapture testing set. \#G: Number of groups for body parts. Skel.: Whether to use a skeleton-aware scanning schema or not. P1: Average MPJPE. \#Params.: Number of parameters. Peak: Peak Memory (MB).}
    \begin{tabular}{ccccc}
      \hline
        \#G & Skel.         & P1 $\downarrow$   & \#Parms. & Peak\\ \hline
        0 & \Checkmark    & 25.6 &  6.8M  & 248.6\\ 
        3 & \Checkmark    & 20.5 &  9.6M  & 265.1\\ 
        5 & \XSolidBrush  & 25.7 & 12.7M  & 290.7\\
        5 & \Checkmark    & \underline{\textbf{17.2}} & 12.3M  & 282.5\\ 
        \hline
    \end{tabular}
\label{tab:group}
\end{table}
\begin{table}[!ht]\small
  \caption{The ablation study for fusion strategies on TotalCapture test set. M: Methods. PM: PoseMamba (vision-only)~\cite{huangPoseMambaMonocular3D2025}, SA: Self-Attention (vision-only with Self-Attention module), CM: Cross-Mamba (visual-inertial Mamba-based fusion), CA: Cross-Attention (visual-inertial fusion with Cross-Attention module). P1: Average MPJPE.} 
    \begin{tabularx}{\linewidth}{ccccccccX}
      \hline
        M &  SW2 & SFS3 & SA3 & UW2 & UFS3 & UA3 & P1 $\downarrow$  \\ \hline
        PM &15.6 & 38.4 & 16.4 & 38.0 & 62.9 & 53.3 & 28.1 \\ 
        SA &15.4 & 35.5 & 17.1& 41.2& 58.8 &51.7 & 26.9 \\ 
        CM &12.3 & 26.3 & 15.4& 31.2& 48.0 &41.9 & 24.3 \\ 
        CA &\underline{\textbf{8.0}} & \underline{\textbf{23.7}} & \underline{\textbf{14.2}} & \underline{\textbf{25.7}} & \underline{\textbf{42.3}} & \underline{\textbf{36.3}} & \underline{\textbf{17.2}} \\ \hline
    \end{tabularx}
\label{tab:ablation}
\end{table}
\subsection{Ablation Studies}
\subsubsection{Stability of Variable-Length Training Strategy}
To evaluate the robustness of the variable-length training strategy, we compare the performance using fixed-length inputs (9, 27, and 81 frames) against a dynamic approach, where the sequence lengths are randomly sampled from ${9, 18, \dots, 81}$ during training. Table~\ref{tab:dynamic} presents the results with the ground-truth 2D keypoints, assessed via P1. 
The variable-length configuration (denoted as V) yields 18.9 mm on P1, approaching the optimal fixed-length result (17.2 mm at $T=81$). Importantly, this approach maintains consistent accuracy across varying lengths, ensuring stability while accommodating flexible sequence processing.

\subsubsection{Effectiveness of Feature Extractor}
To evaluate the contributions of part-aware and skeleton-aware feature extraction, we conduct an ablation study on TotalCapture dataset using the ground-truth 2D keypoints. 
As detailed in Table~\ref{tab:group}, applying STMamba directly to ungrouped visual and inertial inputs (\#G = 0) reduces the model size to 6.8 million parameters and peak memory to 248.6 MB, yet incurs a substantial accuracy degradation of 25.6 mm MPJPE.
Introducing a three-part segmentation (trunk, upper limbs, lower limbs; see appendix for details) improves performance by over 5 mm, reducing the error to 20.5 mm. 
This gain underscores the benefit of the grouping method in enhancing robustness. 
Further increasing the number of groups to five yields an additional improvement, achieving the lowest error of 17.2 mm, while maintaining a moderate increase in model parameters. 
This progression confirms that finer anatomical decomposition improves the quality of estimation.
Moreover, removing the skeleton-aware mechanism and relying solely on the part-aware STMamba for visual feature extraction leads to a sharp accuracy degradation (25.7 mm). 
These findings validate the essential integration of kinematic priors via skeleton-aware modeling and underscore the synergistic benefits of anatomical grouping and modality-specific scanning in VIMCAN.

 \subsubsection{Effectiveness of Multimodal Inputs}
To evaluate the contribution of multimodal inputs, we compare VIMCAN against a vision-only PoseMamba (PM)~\cite{huangPoseMambaMonocular3D2025}, and then substitute the Cross-Attention module with Self-Attention module for visual inputs while maintaining the same grouping strategy (SA). 
As summarized in Table~\ref{tab:ablation}, incorporating Self-Attention module in PoseMamba reduces the P1 from 28.1 mm to 26.9 mm, a modest improvement of 1.2 mm. 
In contrast, VIMCAN's Cross-Attention integration of inertial data (CA) yields 17.2 mm on P1, a substantial 9.7 mm improvement over the vision-only Self-Attention baseline.

\subsubsection{Performance Comparison of Fusion Modules}
To demonstrate VIMCAN's effectiveness in addressing fine-grained cross-modal spatial reasoning, we ablate the fusion mechanism by comparing the integrated Cross-Attention module with a Cross-Mamba alternative. 
Both modules utilize identical IMUs and hyperparameters, with performance evaluated on fixed-length ($T$=81) and ground-truth keypoints. 
As reported in Table~\ref{tab:ablation}, the Cross-Attention module (CA) consistently outperforms the Cross-Mamba module (CM), particularly on unseen subject testing sets (e.g., UFS3 and UA3), where P1 improvements exceed 5.6 mm. The P1 performance for Cross-Attention (17.2 mm) surpasses Cross-Mamba (24.3 mm), underscoring the hybrid architecture’s ability to model complex cross-modal spatial dependencies. 

\subsubsection{The Impact of Mamba Hyperparameters}
To assess the influence of Mamba hyperparameters, we ablate the embedding dimension ($D_e$) and group dimension ($D_g$).
Experiments employ SimpleNet-derived 2D poses with fixed sequence length ($T$=27) on the TotalCapture testing set, measuring MPJPE, parameter count, and throughput (FPS) on an NVIDIA RTX 4060 Laptop GPU.
Results in Table~\ref{tab:hyperparams} reveal distinct trends. Configurations 1--3 show that increasing $D_g$ from 64 to 256 (with $D_e=64$) markedly enhances accuracy, reducing P1 from 36.7 mm to 31.2 mm, albeit at the expense of parameter count and FPS. 
Configuration 2 offers a strong balance of efficiency and accuracy, making it ideal for resource-constrained deployments. 
Configuration 3 delivers the lowest P1 (31.2 mm) while sustaining over 60 FPS, and is thus adopted as the primary setting.
In contrast, raising $D_e$ from 64 to 96 (Configuration 3 and 4) degrades performance, likely due to overfitting.
Furthermore, configurations 1, 5, and 6 enforce $D_e = D_g$, achieving higher throughput (FPS $>$ 70) and lower computational cost (less parameters), but fail to match the best accuracy of the asymmetric configuration.
These findings guide the selection of hyperparameters for practical deployment scenarios.
\begin{table}[!t]\small
  \vspace{-0.2cm}
  \centering
  \caption{The ablation study for model hyperparameters. No.: Configuration index. $D_{e}$: Embedding dimension. $D_{g}$: Group dimension. P1: Average MPJPE. \#Params.: Number of parameters. FPS: Frames per second.}
    \begin{tabular}{cccccc}
      \hline
        No.& $D_{e}$ & $D_{g}$ & P1 $\downarrow$ & \#Parms. & FPS $\uparrow$\\ \hline
        1 & 64 &  64 & 36.7&  1.8M &\underline{\textbf{74.9}}\\
        2 & 64 & 128 & 34.5&  3.9M &73.3\\
        3 & 64 & 256 & \underline{\textbf{31.2}}& 12.3M &61.4\\
        4 & 96 & 256 & 35.2& 13.7M &60.3\\
        5 & 96 &  96 & 35.6&  4.1M &72.7\\ 
        6 &128 & 128 & 34.5&  7.1M &71.1\\  \hline
    \end{tabular}
\label{tab:hyperparams}
\vspace{-0.3cm}
\end{table}
\begin{table}[!t]
  \centering
  \caption{The comparison for computational efficiency on TotalCapture testing set. P1: Average MPJPE (mm). \#Params.: Number of parameters. Peak: Peak Memory (MB). FPS: Frames Per Second.} 
    \begin{tabular}{ccccc}
      \hline
        Methods & P1 $\downarrow$ & \#Params. & Peak $\downarrow$ & FPS $\uparrow$  \\ \hline
        Wang’s~\cite{wangDeepLearningenabledVisualinertial2025} & 34.9 & 7.3M & 969.8 & 45.8\\ 
        CrossMamba & 35.3 & 12.5M & \underline{\textbf{89.6}}& 64.6\\ 
        VIMCAN-B & \underline{\textbf{31.2}} & 12.3M & 282.5& 61.4\\
        VIMCAN-T & 34.5 & 3.9M & 156.4& \underline{\textbf{71.1}} \\ \hline
    \end{tabular}
\label{tab:perf-analysis}
\vspace{-0.3cm}
\end{table}

\subsection{Performance Analysis}
As demonstrated in Table~\ref{tab:perf-analysis}, we compare VIMCAN with its Cross-Mamba variant and Wang’s model (GCN-Transformer)~\cite{wangDeepLearningenabledVisualinertial2025} on consumer-grade hardware (NVIDIA RTX 4060 Laptop), evaluating accuracy (P1), model size (\#Params.), peak memory usage (Peak), and inference speed (FPS).
VIMCAN-B (Balance) corresponds to Configuration 3 in Table~\ref{tab:hyperparams}, while VIMCAN-T (Tiny) corresponds to Configuration 2. 
Since Wang’s~\cite{wangDeepLearningenabledVisualinertial2025} code is not publicly available, we reimplemented their model, achieving a comparable parameter count (7.3M vs. 7.9M reported) while maintaining a reasonable computational scale.
VIMCAN-B achieves the lowest MPJPE (31.2 mm), outperforming Wang’s~\cite{wangDeepLearningenabledVisualinertial2025} (34.9 mm) and Cross-Mamba (35.3 mm), with a peak memory usage of 282.5 MB and FPS of 61.4. 
VIMCAN-T trades slight accuracy (34.5 mm) for a reduced model size (3.9M) and higher FPS (71.1), with a lower peak memory usage of 156.4 MB. This trade-off, as discussed above, makes VIMCAN-B preferable for accuracy-critical applications, while VIMCAN-T is suitable for resource-constrained deployments.
\section{Conclusion}
In this work, we introduce VIMCAN, a hybrid visual-inertial Mamba-Cross-Attention network for 3D HPE. 
Our approach integrates RGB keypoints and IMU data via Mamba's efficient sequence modeling and Cross-Attention's robust multimodal fusion, mitigating depth ambiguity.
Extensive experiments demonstrate VIMCAN's efficacy. It significantly outperforms a leading GCN-Transformer model on TotalCapture, reducing MPJPE by 11.4 mm while requiring only 29\% of its memory footprint and delivering 1.3$\times$ throughput. Furthermore, VIMCAN achieves a competitive 45.3 mm MPJPE on the challenging 3DPW dataset.
Consequently, it achieves a better balance of accuracy and efficiency than prior RNN, CNN, and Transformer-based methods. VIMCAN natively supports variable-length sequences, making it well-suited for real-time applications in motion capture, rehabilitation, and human-computer interaction. One limitation is its reliance on strict sensor calibration. Future work will explore adaptive alignment techniques to enhance robustness.

\noindent
\textbf{Acknowledgments.}
This work was supported by 
the National Key R\&D Program of China (No. 2022ZD0115902), 
the National Science and Technology Major Project (No. 2025ZD1605703), 
the National Natural Science Foundation of China (No. 32571260), 
the Open Project Program of State Key Laboratory of Virtual Reality Technology and Systems, Beihang University (No. VRLAB2024C06), 
and the Fundamental Research Funds for Beijing Municipal Universities (Grant No. 154225001). 
The authors gratefully acknowledge the financial support.

{
    \small
    \bibliographystyle{ieeenat_fullname}
    \bibliography{main}
}


\end{document}



\clearpage
\setcounter{page}{1}
\maketitlesupplementary

\section{Implementation of Cross-Attention}
\begin{figure}[!ht]
    \centering
    \includegraphics[width=0.7\linewidth]{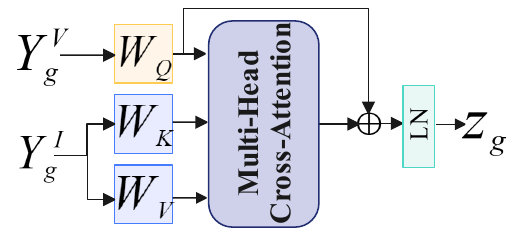}
    \caption{The architecture of the Cross-Attention module.}
    \label{fig:ca}
    \vspace{-0.3cm}
\end{figure}

As illustrated in Fig.~\ref{fig:ca}, we adopt a standard Cross-Attention mechanism to integrate multimodal features. For each group $g$, visual features $Y^V_g$ are projected into queries $Q^V_g$, while inertial features $Y^I_g$ are projected into keys $K^I_g$ and values $V^I_g$ via separate linear layers. 
To preserve the skeletal structural information, a residual connection is applied exclusively to the visual queries. 

\section{Group Partition Strategy}
\label{sec:group_partition}
As shown in Table~\ref{tab:group_partition}, Group 0 uses all available keypoints and IMUs without any partitioning. Group 3 divides the human body into three regions and Group 5 further refines the human body into five parts.

This hierarchical partitioning facilitates more structural and modality-specific feature extraction, which is essential for robust human pose estimation.

\begin{table}[h]\small
\centering
\caption{The group configurations and their corresponding components for visual keypoints and inertial sensors. The indice is presented in Fig.3 of the main paper.}
\begin{tabular}{|c|c|c|c|}\hline
Groups & Partition & Visual & Inertial \\ \hline
   0   &     -     & All keypoints & All IMUs \\ \hline
 \multirow{3}{*}{3}&Torso & $V_{0,7,8,9,10}$ & $I_{0,1}$ \\ 
                &Upper & $V_{11,12,13,14,15,16}$ & $I_{4,5}$ \\ 
                &Lower & $V_{1,2,3,4,5,6}$ & $I_{2,3}$ \\ \hline
 \multirow{5}{*}{5}&Torso & $V_{0,7,8,9,10}$ & $I_{0,1}$ \\ 
                &Left Arm & $V_{0,7,8,11,12,13}$ & $I_{0,1,4}$ \\ 
                &Rigth Arm & $V_{0,7,8,14,15,16}$ & $I_{0,1,5}$ \\ 
                &Left Leg & $V_{0,4,5,6}$ & $I_{0,2}$ \\ 
                &Rigth Leg & $V_{0,1,2,3}$ & $I_{0,3}$ \\ \hline
\end{tabular}
\label{tab:group_partition}
\vspace{-0.3cm}
\end{table}

\section{Data Preprocessing for 2D Keypoints}
\label{sec:data_preprocess_2d}
The input keypoints are derived from 2D detectors (\textit{i.e.}, MediaPipe and SimpleNet) and mapped to a predefined set of $J=17$ body joints. 
For joints with direct correspondences (\textit{e.g.}, LeftShoulder, RightKnee), we use the respective landmark coordinates. For composite joints such as Hip, Spine, Spine3, and Neck, we apply geometric computations to approximate their positions:
\begin{itemize}
\item \textbf{Hips}: Computed as the midpoint between the left and right hip landmarks.
\item \textbf{Spine}: Interpolated 25\% of the way from the hip center to the shoulder center.
\item \textbf{Spine3}: Interpolated 75\% of the way from the hip center to the shoulder center.
\item \textbf{Neck}: Defined as the point 33\% of the way from the shoulder center to the nose.
\end{itemize}

To achieve scale and translation invariance, we normalize the 2D keypoints as follows. First, we compute the bounding box scale as the maximum of the width and the height of the keypoint set. Each keypoint is then divided by this scale and is represented in the form of root-relative coordinates.

\section{Impact of 2D Detector}
\label{sec:2d_detector}
To quantitatively assess the influence of upstream 2D pose detection on our framework, we compare MediaPipe and SimpleNet on TotalCapture testing set. 
Table~\ref{tab:2d_per_joint} reports their per-joint and overall performance in terms of MPJPE (in pixels) and Percentage of Correct Keypoints (PCK) at two thresholds (\textit{i.e.}, 25, 50).
SimpleNet achieves a lower MPJPE compared to MediaPipe, indicating higher detection accuracy. This performance gap is further reflected in the PCK metrics, where SimpleNet consistently outperforms MediaPipe at both thresholds.
These improvements in the quality of 2D detection directly contribute to the enhanced performance of 3D pose estimation, as observed in Table 1 of the main paper.
This analysis confirms that while our framework demonstrates robustness to varying levels of 2D detection noises, the overall system performance benefits significantly from higher-quality upstream detection.

\begin{table}[ht]\small
\centering
\caption{The comparison of the per-joint and overall performance for 2D pose detectors on TotalCapture testing set. MP: MediaPipe. SN: SimpleNet. MPJPE: Average MPJPE (pixel, lower is better). PCK: Percentage of Correct Keypoints (percentage, higher is better). 50 and 25 are thresholds.}
\begin{tabular}{c|cc|cc|cc}
\hline
\multirow{2}{*}{Joint} & \multicolumn{2}{c|}{MPJPE $\downarrow$} & \multicolumn{2}{c|}{PCK@50 $\uparrow$} & \multicolumn{2}{c}{PCK@25 $\uparrow$} \\
& MP & SN & MP & SN & MP & SN \\
\hline
Hips & 33.2 & \textbf{16.2} & 98.6 & 98.6 & 86.2 & \textbf{90.1} \\
RightUpLeg & 32.9 & \textbf{18.1} & 98.0 & \textbf{98.1} & 85.2 & \textbf{89.1} \\
RightLeg & 35.5 & \textbf{17.9} & 96.4 & \textbf{97.4} & 87.2 & \textbf{90.7} \\
RightFoot & 37.8 & \textbf{20.2} & 94.9 & \textbf{95.7} & 86.8 & \textbf{90.5} \\
LeftUpLeg & 34.3 & \textbf{19.6} & 98.4 & 98.4 & 84.8 & \textbf{86.1} \\
LeftLeg & 32.8 & \textbf{15.1} & 96.6 & \textbf{97.6} & 88.2 & \textbf{92.8} \\
LeftFoot & 35.2 & \textbf{17.2} & 95.3 & \textbf{96.2} & 88.5 & \textbf{92.2} \\
Spine & 39.5 & \textbf{25.1} & 97.4 & 97.4 & 75.7 & \textbf{77.1} \\
Spine3 & 38.1 & \textbf{19.3} & 94.2 & \textbf{96.6} & 80.8 & \textbf{87.4} \\
Neck & 41.0 & \textbf{25.4} & 96.1 & 96.1 & 64.8 & \textbf{66.7} \\
Head & 34.8 & \textbf{19.5} & 98.7 & \textbf{98.8} & 85.9 & \textbf{86.9} \\
LeftArm & 44.7 & \textbf{32.7} & 94.0 & \textbf{98.8} & 40.4 & \textbf{49.2} \\
LeftForeArm & 35.8 & \textbf{20.3} & 98.1 & \textbf{98.2} & 78.7 & 78.7 \\
LeftHand & 37.6 & \textbf{19.0} & 94.5 & \textbf{95.9} & 76.7 & \textbf{86.1} \\
RightArm & 40.3 & \textbf{21.1} & 91.1 & \textbf{93.8} & 74.4 & \textbf{85.2} \\
RightForeArm & 34.9 & \textbf{19.0} & 97.6 & \textbf{98.2} & 85.7 & \textbf{85.9} \\
RightHand & 37.8 & \textbf{18.8} & 95.7 & \textbf{97.5} & 81.5 & \textbf{88.7} \\ \hline
Overall & 36.9 & \textbf{20.3} & 96.2 & \textbf{96.7} & 80.5 & \textbf{82.1} \\ 
\hline
\end{tabular}
\label{tab:2d_per_joint}
\vspace{-0.3cm}
\end{table}